\documentclass[pmlr,twocolumn,10pt]{jmlr} 

\mlhtrack{proceedings}

\newif\iffinal
\finaltrue  

\iffinal
    \ifmlhneedspmlr
      \jmlrvolume{346}
      \jmlryear{2026}
    \fi
    \ifmlhfindings \jmlrproceedings{}{ML4H 2026 - Findings Track}\fi
    \ifmlhdemo     \jmlrproceedings{}{ML4H 2026 - Demo Track}\fi
    \jmlrworkshop{Machine Learning for Health (ML4H) 2026}
\else
    \jmlrproceedings{}{Submitted to ML4H 2026: \mlhtrackname}
    \jmlrworkshop{Machine Learning for Health (ML4H) 2026}
\fi

\usepackage{booktabs}
\usepackage{float} 
\usepackage{siunitx}

\usepackage[switch]{lineno}

\theorembodyfont{\upshape}
\theoremheaderfont{\scshape}
\theorempostheader{:}
\theoremsep{\newline}

\title[Target Recoverability under Selective Labels]
{Target Recoverability under Selective Labels: A Failure Mode in ECG-Based Acute Coronary Syndrome Prediction}

\author{%
  \Name{Alexander Schubert}
  \Email{alexander\_schubert@berkeley.edu}\\
  \addr UC Berkeley and UCSF
  \AND
  \Name{Anton H{\aa}kansson}
  \Email{anton.hakansson@med.lu.se}\\
  \addr Lund University 
  \AND
  \Name{Ziad Obermeyer}
  \Email{zobermeyer@berkeley.edu}\\
  \addr UC Berkeley
}

\begin{document}

\maketitle

\ifmlhdemo\else

\begin{abstract}
Medical machine learning models often learn from diagnoses observed only in patients clinicians chose to test. The biases of such selective labels are well known, but less understood is what happens when a diagnosis includes cases the model’s inputs cannot detect. Because testing decisions draw on information the model does not see, these cases can appear predictable among tested patients, producing relationships that fail, and can even reverse, among the untested. We formalize this mechanism and show in simulations that it worsens as testing becomes more selective or undetectable cases more common, while models trained only on detectable cases remain informative. We empirically study electrocardiogram (ECG)-based prediction of acute coronary syndrome (ACS), which is confirmed only by invasive testing and whose subtypes differ in how clearly they appear on the ECG. Three ECG foundation models fine-tuned on revascularization, our ACS proxy, performed well among catheterized patients (AUROC 0.740--0.770) but worse than chance for 30-day major adverse cardiovascular events among untested patients (AUROC 0.381--0.478). Models trained on ECG-visible subtypes retained discrimination (AUROC 0.680--0.709), whereas a model trained on unstable angina, which can occur without ECG changes, was inversely related to risk (AUROC 0.310). Under selective labels, effective learning therefore requires prediction targets that can be inferred from the model’s inputs and evaluation in the intended deployment population, which often extends beyond the patients who were tested.
\end{abstract}


\fi

\ifmlhneedsstatements

\paragraph*{Data and Code Availability}
This study uses two publicly available, de-identified ECG datasets, described in Section~\ref{sec:data}. Code is available on \href{https://github.com/alexmschubert/selective_labels_recoverability}{GitHub}.

\paragraph*{Institutional Review Board (IRB)}
This study uses only publicly available, de-identified data. The secondary analyses presented here do not require additional Institutional Review Board approval.


\fi

\section{Introduction}
\label{sec:intro}

Clinical machine learning models are often trained on labels that are not measured independently of care, but are revealed only after clinicians decide whom to test. Disease status may become known only after angiography, biopsy, imaging, or laboratory testing, while outcomes remain unobserved for patients who do not undergo the corresponding diagnostic procedure. This creates a \emph{selective labels} problem: because testing decisions depend on observed and unobserved clinical information, patients with observed labels need not be representative of the population in which a model will ultimately be used \citep{ransohoff1978spectrum,begg1983verification,lakkaraju2017selective}. In clinical diagnostics, the closely related phenomenon of \emph{partial verification bias} has long been recognized when application of a definitive reference standard depends on earlier test results or clinical judgment \citep{kohn2022partial,arifin2022verification}. Recent work has shown that such selection can distort both model development and evaluation. Differences in testing can induce biased training labels and disparities across patient groups \citep{chang2022disparate,chang2024selective}, while performance estimated only among patients with observed labels can differ substantially from performance in the full deployment population \citep{corbin2023labelselection}. 

A critical but less examined question in learning under selective labels is whether the target is recoverable from the model input, and how this recoverability interacts with selection for testing. A clinically valid outcome may contain components for which the input modality carries little or no predictive information. In the full population, such components are simply difficult to predict. Under selective labels, however, this need not remain true. If selection depends both on features represented in the model input and on additional clinical information related to the outcome, conditioning on being selected can induce associations between the input and otherwise poorly recoverable outcome components, a form of selection or collider bias \citep{cole2010collider,weiskopf2023collider}. A model trained only in the selected population can therefore appear to predict these components by exploiting relationships created by the selection process, despite the absence of genuine physiological signals. We refer to this interaction between target recoverability and selective labels as a potential source of transport failure. 
Figure~\ref{fig:selection_mechanism} illustrates how selective testing can make otherwise normal-looking inputs appear predictive of disease among labeled patients.

\begin{figure*}[!t]
\floatconts
{fig:selection_mechanism}
{\caption{
\textbf{Schematic depiction of the bias mechanism.}
Testing depends on both the model input and additional clinical information.
When this additional information identifies disease not detectable from
the input, inputs without disease signal can become associated with higher risk among labeled patients, even though this relationship does not hold
among unlabeled patients.
}}
{
\centering
\includegraphics[width=\textwidth]{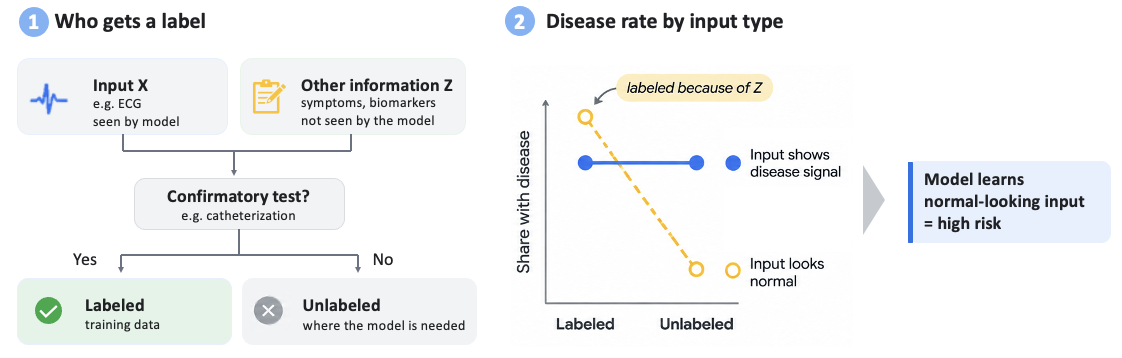}
}
\end{figure*}

Acute coronary syndrome (ACS) provides a clinically important setting in which selective labels and target recoverability intersect. ACS encompasses several forms of acute myocardial ischemia, and the 12-lead electrocardiogram (ECG) is central to initial diagnosis and triage \citep{rao2025acs}. Crucially, however, ACS phenotypes differ in how clearly their underlying pathology manifests on the ECG: some produce characteristic ischemic patterns, whereas others may occur despite a normal or nondiagnostic ECG \citep{rao2025acs,gulati2021chestpain}. At the same time, definitive disease labels are only observed  among patients selected for invasive angiography. This combination of heterogeneous ECG recoverability and selective outcome ascertainment makes ACS a natural setting in which to study their interaction.

In this work, we develop a framework for understanding how target recoverability interacts with selective labels. We formalize a mechanism by which conditioning on selection can make outcome components that are weakly recoverable from the model input appear predictive within the labeled population, causing models to exploit selection-induced associations. We characterize this mechanism theoretically and in controlled synthetic experiments that vary both target recoverability and the strength of selection. We then examine the phenomenon empirically in ACS using two independent ECG cohorts, comparing models trained on alternative clinical targets, from phenotypes with strong ECG manifestations, such as STEMI and OMI, to phenotypes that may be less directly recoverable from the ECG, such as unstable angina (UA). We then evaluate how their predictive relationships change between selectively tested and broader untested populations. These analyses present target recoverability as an important dimension to consider when designing and evaluating learning problems under selective labeling.

\section{Background}
\label{sec:background}

\subsection{Selective Labels}
\label{sec:selective_labels}

Selective labels arise when the availability of an outcome depends on a prior decision, so that labels are observed for only a non-random subset of the population. The problem was formalized in machine learning by \citet{lakkaraju2017selective}, who showed that evaluating predictions against outcomes generated by existing human decisions can fundamentally limit what can be learned about alternative decisions. In diagnostic medicine, the closely related problem has long been studied as \emph{verification bias}: estimates of diagnostic accuracy can be biased when definitive disease status is established only for patients selected to receive a reference-standard test \citep{begg1983verification}. This problem is particularly relevant to clinical machine learning because diagnostic testing itself is driven by patient characteristics, clinician judgment, and resource availability. \citet{chang2022disparate} showed that differential testing can generate systematically biased training labels, while \citet{corbin2023labelselection} demonstrated that restricting evaluation to patients with observed labels can substantially bias estimates of model performance in the intended deployment population. Existing work therefore establishes that the mechanism governing \emph{which} labels are observed is central to both model development and evaluation; here, we additionally study how that mechanism interacts with \emph{what information about the target is recoverable from the model input}.

\subsection{Target Choice and Distribution Shift}
\label{sec:targets_shift}

The choice of prediction target determines what signal a model is optimized to learn. In clinical data, the desired biological or clinical quantity is often unavailable and replaced by an observable proxy; when that proxy is systematically influenced by healthcare access, utilization, measurement, or treatment, a model can predict it accurately while learning a distorted representation of the underlying clinical construct, leading to biased downstream decisions \citep{obermeyer2019dissecting,mullainathan2021predicting,tal2023target}. A distinct but equally important failure mode arises from \emph{distribution shift} between model development and deployment. Changes in patient populations and disease prevalence, clinical practice and treatment, measurement technologies, documentation, or other aspects of the healthcare system can alter the statistical relationships on which a model relies \citep{subbaswamy2020deployment,finlayson2021clinician}. Clinical models may consequently exploit associations that are highly predictive within the development environment but are not transportable \citep{zech2018generalization, degrave2021shortcuts, guo2023ehrshift}. We study a related mechanism in which selective observation itself changes the apparent relationship between the model input and the prediction target, making a weakly recoverable target component appear predictive within the selected population.

\subsection{Acute Coronary Syndrome}
\label{sec:acs_background}

Acute coronary syndrome (ACS) describes acute myocardial ischemia resulting from a sudden reduction in coronary blood flow and conventionally encompasses unstable angina (UA), non-ST-segment elevation myocardial infarction (NSTEMI), and ST-segment elevation myocardial infarction (STEMI) \citep{rao2025acs,byrne2023esc}. Diagnosis integrates symptoms, the 12-lead electrocardiogram (ECG), and cardiac troponin, but ACS phenotypes differ substantially in how directly they manifest on the ECG. STEMI is defined by characteristic ST-segment abnormalities, while the newer occlusion myocardial infarction (OMI) paradigm identifies additional patterns associated with acute coronary occlusion even when conventional STEMI criteria are not met \citep{meyers2021omi}. In contrast, electrocardiographic abnormalities are not required for non-ST-elevation ACS: NSTEMI and UA may present with nonspecific changes or a normal ECG, and a normal ECG does not exclude ACS \citep{rao2025acs,gulati2021chestpain}. UA is characterized by myocardial ischemia without troponin-detectable myonecrosis, making its diagnosis particularly dependent on symptoms and additional clinical testing beyond a single ECG alone \citep{rao2025acs}. Consequently, a broad ACS-related target can combine phenotypes that differ markedly in their recoverability from the ECG. Definitive characterization of coronary anatomy and identification of lesions leading to revascularization require invasive coronary angiography, which is performed only in patients selected for invasive evaluation. This creates the selective-label setting studied here: detailed coronary outcomes are available in catheterized patients but absent in the much larger population that is not catheterized.

\section{Data}
\label{sec:data}

We use two complementary publicly available ECG datasets: the Chongqing angiography cohort and the Nightingale BWH emergency department cohort. Both contain ECG waveforms linked to clinically meaningful coronary outcomes, but differ in a way that is central to our analysis. In Chongqing, all patients underwent coronary angiography, providing detailed labels for alternative ACS phenotypes within a selected population. In Nightingale, ECGs are available for patients both inside and outside the subsequent testing pathway, allowing us to compare predictive relationships between selectively tested and untested populations.

\subsection{Chongqing Angiography Cohort}
\label{sec:chongqing_data}

We use the publicly available ECG dataset from the First Affiliated Hospital of Chongqing Medical University \citep{du2026acs}. The dataset contains 19,955 ten-second, 12-lead ECGs sampled at 500\,Hz from 18,909 patients who underwent coronary angiography between 2015 and 2024; only ECGs recorded before angiography are included \citep{du2026acs}. Clinical annotations include ST-segment elevation myocardial infarction (STEMI), non-ST-segment elevation myocardial infarction (NSTEMI), unstable angina (UA), acute myocardial infarction (AMI), and occlusion myocardial infarction (OMI). OMI is defined as an acute culprit-vessel occlusion with TIMI flow grade 0 or 1 before percutaneous coronary intervention. We additionally define broad acute coronary syndrome (ACS) as STEMI, NSTEMI, or UA. For our analysis, we require a corresponding waveform file with the expected 5,000 samples and exclude malformed or incomplete waveform records. Detailed cohort characteristics and label prevalences are reported in Appendix~\ref{app:chongqing_data}.

\subsection{Nightingale Emergency Department Cohort}
\label{sec:nightingale_data}

We use the Nightingale Open Science \emph{Assessing Heart Attack Risk} dataset from an urban academic hospital \citep{anderson2023bwh}. The dataset links ECGs obtained during emergency-department encounters to subsequent cardiac testing, coronary revascularization, and 30-day clinical outcomes. We classify encounters as \emph{catheterized} if cardiac catheterization occurred within 10 days and \emph{untested} if neither catheterization nor a stress test procedure was ordered for the patient. Among catheterized encounters, revascularization is defined as stent placement or coronary artery bypass grafting (CABG) within 10 days. Across all encounters, we retain 30-day major adverse cardiovascular events (MACE), troponin-positive MACE, and death as outcome measures. We apply the dataset's predefined modeling exclusions and retain all qualifying ECGs with observed outcomes for the primary analyses. Detailed cohort definitions and summary statistics are reported in Appendix~\ref{app:nightingale_data}.


\section{Bias from Partial Recoverability under Selective Labels}
\label{sec:methods}

\subsection{Setting and Failure Mechanism}
\label{sec:bias_mechanism}

Let $X$ denote the model input (an ECG in our clinical
application), $S\in\{0,1\}$ indicate whether a patient is
selected for definitive testing, and $Y$ denote the clinical target. The
definitive label is observed only when $S=1$. We consider a target composed
of two positive mechanisms,
\begin{equation}
    Y = Y_R \lor Y_U,
\end{equation}
where $Y_R$ denotes a component that is recoverable from $X$ and $Y_U$ a
component that is weakly recoverable or unrecoverable from $X$. For clarity,
consider the idealized case
\begin{equation}
\begin{aligned}
    P(Y_R=1\mid X=x) &= p_R(x), \\
    P(Y_U=1\mid X=x) &= \pi_U.
\end{aligned}
\end{equation}
so that $p_R(x)$ varies with the input, whereas $Y_U$ contains no predictive
signal in $X$.

Under selective labels, however, training does not recover
$P(Y=1\mid X=x)$. A sufficiently flexible classifier trained only on tested
patients instead approaches
\begin{equation}
    f_{\mathrm{sel}}(x)
    =P(Y=1\mid X=x,S=1),
\end{equation}
whereas application to patients who have not yet been selected for testing
depends on
\begin{equation}
    f_{\mathrm{unt}}(x)
    =P(Y=1\mid X=x,S=0).
\end{equation}

The distinction becomes important when evidence outside $X$ both predicts
$Y_U$ and influences testing. Let

\begin{equation}
\begin{aligned}
a(x) &= P(S=1 \mid X=x, Y_U=1), \\
b(x) &= P(S=1 \mid X=x, Y_U=0).
\end{aligned}
\end{equation}

By Bayes' rule,
\begin{equation}
\begin{aligned}
\frac{P(Y_U=1 \mid x,S=1)}
     {P(Y_U=0 \mid x,S=1)}
&=
\frac{\pi_U}{1-\pi_U} \\
&\quad \times \frac{a(x)}{b(x)}.
\end{aligned}
\label{eq:selected_enrichment}
\end{equation}

while among untested patients,
\begin{equation}
\begin{aligned}
\frac{P(Y_U=1 \mid x,S=0)}
     {P(Y_U=0 \mid x,S=0)}
&=
\frac{\pi_U}{1-\pi_U} \\
&\quad \times
\frac{1-a(x)}{1-b(x)}.
\end{aligned}
\label{eq:untested_depletion}
\end{equation}

Equations~\ref{eq:selected_enrichment}--\ref{eq:untested_depletion}
expose the failure mode. Consider an ECG $x_L$ for which
$p_R(x_L)$ is small, indicating a low probability of the
recoverable disease component. In clinical practice, a patient
with such an ECG may nevertheless undergo definitive testing
because of concerning symptoms, elevated biomarkers, or other
information unavailable to the ECG model. If testing is
uncommon among patients with this ECG in the absence of $Y_U$,
then $b(x_L)$ is small. If $Y_U$ is associated with additional
clinical findings that prompt testing, then $a(x_L)>b(x_L)$.
Consequently, among patients with the same ECG $x_L$, those
selected for testing have a higher prevalence of $Y_U$ than
the underlying population, whereas those not selected have
a lower prevalence. Specifically,
\begin{equation}
\begin{aligned}
P(Y_U=1 \mid x_L,S=1)
&> \pi_U \\
&> P(Y_U=1 \mid x_L,S=0).
\end{aligned}
\end{equation}
Thus, conditioning on testing creates different associations
between the ECG and $Y_U$, even though $Y_U$ is independent
of $X$ in the underlying population.

This is a selection-induced association: conditioning on testing opens a path
between information contained in $X$ and extra-input clinical information
that also drives testing, a form of collider bias
\citep{cole2010collider, weiskopf2023collider}. A model trained in the selected population can therefore
learn to assign higher risk to inputs with low $p_R(x)$,
reflecting their association with $Y_U$ among tested patients.

To see when this affects ranking, suppose $Y_R$ and $Y_U$ are mutually
exclusive and the relationship between $X$ and the recoverable component is
otherwise stable across selection. Then approximately
\begin{equation}
\begin{aligned}
    f_{\mathrm{sel}}(x) &= p_R(x)+p_U^{(1)}(x),\\
    f_{\mathrm{unt}}(x) &= p_R(x)+p_U^{(0)}(x).
\end{aligned}
\end{equation}
where $p_U^{(s)}(x)=P(Y_U=1\mid X=x,S=s)$. For a low-risk-appearing input
$x_L$ and a high-risk input $x_H$ with
$p_R(x_L)<p_R(x_H)$, the selected-sample ranking reverses the
recoverable signal whenever

\begin{equation}
p_U^{(1)}(x_L)-p_U^{(1)}(x_H)
>
p_R(x_H)-p_R(x_L).
\label{eq:rank_reversal}
\end{equation}
In contrast, among untested patients, $Y_U$ is less prevalent than in the underlying population when $a(x)>b(x)$. Their risk ordering can therefore remain
$f_{\mathrm{unt}}(x_H)>f_{\mathrm{unt}}(x_L)$,
opposite to the ordering learned among tested patients.
Selective labels can thus transform lack of recoverability from a simple loss of predictive
information into a systematically misleading predictive relationship.

\subsection{Synthetic Evaluation}
\label{sec:simulation}

We constructed a simulation to isolate the interaction between
partial target recoverability and selective testing. The model-visible signal
was represented by $X\in[0,1]$. We generated mutually exclusive recoverable and
unrecoverable outcome components such that
\begin{equation}
\begin{aligned}
P(Y_R=1\mid X=x,Y_U=0) &= 0.05+0.30x,\\
Y_U &\perp X,
\end{aligned}
\end{equation}
with broad outcome $Y=Y_R\lor Y_U$. We varied target recoverability through
\begin{equation}
q=P(Y_U=1\mid Y=1),
\end{equation}
the fraction of positive broad outcomes attributable to the unrecoverable
component.

Selective testing was induced through an extra-input clinical signal $C$,
which was unavailable to the model but associated with $Y_U$:
$P(C=1\mid Y_U=1)=0.90$ and $P(C=1\mid Y_U=0)=0.10$.
The targeted testing policy was
\begin{equation}
P_{\mathrm{target}}(S=1\mid X=x,C)=
\begin{cases}
0.90, & C=1,\\
0.05+0.85x, & C=0,
\end{cases}
\end{equation}
and selective-testing strength $\lambda\in[0,1]$ interpolated between this
policy and random testing:
\begin{equation}
\begin{aligned}
P_\lambda(S=1\mid X,C)
&=(1-\lambda)0.50\\
&\quad+\lambda P_{\mathrm{target}}(S=1\mid X,C).
\end{aligned}
\end{equation}

Using the simulation's known probabilities, we calculated ideal
tested-population risk scores directly, without fitting models
to sampled data. This isolates the failure mode from estimation
error and shows that it can arise even when a model perfectly
estimates risk among tested patients. Specifically, we calculated
the broad-label and modality-aligned scores,
\begin{equation}
\begin{aligned}
f_{\mathrm{broad}}(x)
    &=P(Y=1\mid X=x,S=1),\\
f_{\mathrm{aligned}}(x)
    &=P(Y_R=1\mid X=x,S=1),
\end{aligned}
\end{equation}
which are optimal probability predictors for their respective
targets in the tested population. We evaluated both scores
in untested patients against the same broad outcome $Y$.
We examined two complementary sweeps: selective-testing strength
($\lambda=0$--$1$) at $q=0.70$, and unrecoverable target share
($q=0$--$0.80$) under fully targeted testing ($\lambda=1$).
Full simulation details are provided in the accompanying
code repository.

\subsection{Empirical Evaluation in Acute Coronary Syndrome}
\label{sec:empirical_framework}

We test the proposed failure mechanism using two complementary empirical
experiments.

\noindent\emph{Learning under selective labels.}
We first study predictors trained directly within the selectively labeled
Nightingale population. Among patients undergoing catheterization, we
fine-tune three ECG foundation models: ST-MEM, ECGFounder, and
ECGFlow \citep{na2024stmem,li2025ecgfounder,moody2025ecgflow} to predict
revascularization by PCI or CABG within 10 days. Revascularization is used as
the observed positive label because coronary anatomy and the subsequent
decision to intervene are available only after patients enter the invasive
testing pathway.

For patients who are not tested, the corresponding coronary label is
unobserved. We therefore evaluate model predictions against 30-day major
adverse cardiovascular events (MACE), an imperfect but clinically meaningful
proxy for potentially missed ACS. Untreated or missed ACS can result in
subsequent myocardial injury, malignant arrhythmia, or death, and 30-day MACE
is routinely used in clinical pathways to quantify short-term risk among
patients evaluated for suspected ACS \citep{gulati2021chestpain,anderson2023bwh}.
Importantly, MACE is not treated as ground-truth ACS: not every missed ACS
produces a MACE event, and MACE can occur for reasons other than missed ACS.
To assess the robustness to the specification of this proxy troponin-positive MACE and 30-day mortality are evaluated as additional outcome definitions.

\noindent\emph{Target recoverability.}
We further isolate the role of target choice using the Chongqing angiography
cohort. Using a common training pipeline, we construct ECG predictors for
broad ACS and for the individual targets OMI, AMI, STEMI, NSTEMI, and UA.
These predictors are then frozen and evaluated externally in Nightingale,
separately among catheterized and untested patients. This comparison holds the
input modality and downstream population fixed while varying the clinical
target used for learning. It therefore tests whether transport across the
selection boundary differs between targets with stronger ECG manifestations,
such as OMI and STEMI, and targets that may be less directly recoverable from
the ECG, such as UA and NSTEMI.

\section{Results}

\begin{figure*}[!t]
\floatconts
{fig:simulation}
{\caption{
\textbf{Simulation of selective testing and target recoverability.}
\textbf{A}, AUC in the untested population as selective-testing strength
increases, with $q=P(Y_U=1\mid Y=1)=0.70$.
\textbf{B}, AUC as the proportion of positive outcomes attributable to the
unrecoverable component increases, with fully targeted testing ($\lambda=1$).
Risk scores are learned among tested patients and evaluated against the same composite outcome $Y$ in untested patients. The dotted line denotes chance discrimination.
}}
{
\includegraphics[width=\textwidth]{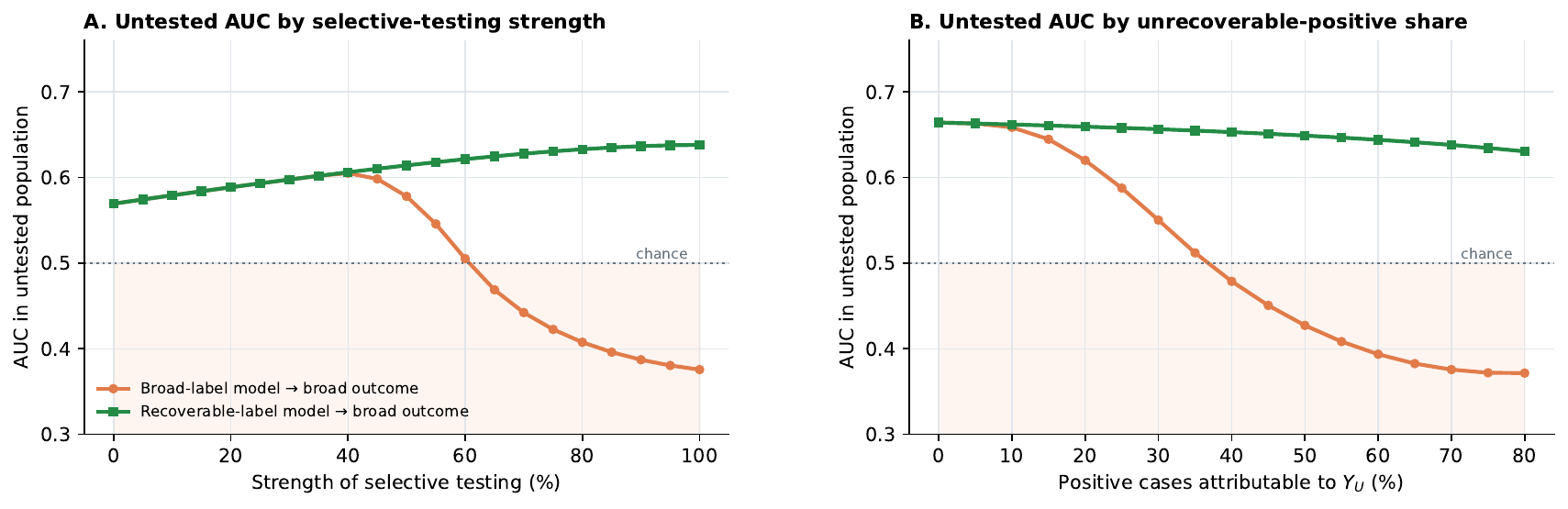}
}
\end{figure*}

\subsection{Synthetic Evaluation}

The synthetic experiment reproduced the predicted failure mode
(Figure~\ref{fig:simulation}). 

 We first varied the testing policy from random testing ($\lambda=0$) to fully targeted testing ($\lambda=1$), holding the proportion of unrecoverable cases fixed at $q=0.70$. Initially, both scores performed similarly, and discrimination among untested patients improved slightly. As testing increasingly selected patients with clinical signs of $Y_U$, fewer such cases remained in the untested population, making the ECG-detectable signal more informative. With stronger selection, however, patients tested despite low-risk ECGs became increasingly more likely to belong to the $Y_U$ group, creating a misleading association in the tested population. The predictions for the broad outcome  $Y$ began to reflect this association. Its AUC fell sharply beyond $\lambda\approx0.4$, dropped below chance near $\lambda=0.6$, and reached 0.38 at $\lambda=1$. The score trained on the recoverable component remained informative.

We next held testing fully targeted ($\lambda=1$) and
varied $q$, the proportion of positive cases attributable
to the unrecoverable component
(Figure~\ref{fig:simulation}B). Performance of the
broad-outcome score declined as these cases became more
common, falling below chance when they accounted for
approximately 35--40\% of positive outcomes. The score
trained on the recoverable component remained above chance
throughout. These results show how selective testing can
turn a poorly recoverable training target into misleading
predictions, even when the model estimates risk correctly
among tested patients.

We also examined whether weighting could correct this
failure (Appendix~\ref{app:ipw}). Inverse-odds weighting
based only on the model input did not prevent the reversal.
Weighting using the true selection probabilities, including
information unavailable to the model, restored positive
discrimination.




\begin{table*}[!tbp]
\floatconts
{tab:transport}
{\caption{
\textbf{ECG model performance in catheterized and untested patients.}
AUCs (95\% CI) for 10-day revascularization among catheterized patients
and 30-day MACE, troponin-positive MACE, and mortality among untested patients.
}}
{
\centering
\scriptsize
\setlength{\tabcolsep}{4pt}
\begin{tabular}{lcccc}
\toprule
\textbf{Model / target} &
\shortstack{\textbf{Cath revasc.}\\\textbf{AUC (95\% CI)}} &
\shortstack{\textbf{Untested MACE}\\\textbf{AUC (95\% CI)}} &
\shortstack{\textbf{Untested trop.-MACE}\\\textbf{AUC (95\% CI)}} &
\shortstack{\textbf{Untested death}\\\textbf{AUC (95\% CI)}} \\
\midrule
\multicolumn{5}{l}{\emph{Trained in Nightingale catheterized cohort}} \\
ST-MEM &
0.770 (0.726--0.812) &
0.478 (0.454--0.495) &
0.497 (0.466--0.520) &
0.442 (0.428--0.457) \\
ECGFounder &
0.745 (0.696--0.787) &
0.422 (0.401--0.437) &
0.437 (0.410--0.457) &
0.396 (0.383--0.410) \\
ECGFlow &
0.740 (0.694--0.782) &
0.381 (0.361--0.395) &
0.385 (0.360--0.403) &
0.387 (0.373--0.400) \\
\midrule
\multicolumn{5}{l}{\emph{Trained in Chongqing angiography cohort}} \\
ACS &
0.687 (0.662--0.710) &
0.517 (0.492--0.535) &
0.515 (0.482--0.538) &
0.523 (0.511--0.535) \\
OMI &
0.636 (0.610--0.660) &
0.709 (0.702--0.716) &
0.712 (0.703--0.720) &
0.682 (0.672--0.692) \\
STEMI &
0.638 (0.612--0.664) &
0.693 (0.683--0.703) &
0.694 (0.680--0.706) &
0.676 (0.666--0.686) \\
NSTEMI &
0.626 (0.602--0.652) &
0.680 (0.672--0.688) &
0.684 (0.673--0.693) &
0.654 (0.644--0.664) \\
UA &
0.486 (0.460--0.512) &
0.310 (0.296--0.321) &
0.307 (0.289--0.321) &
0.332 (0.322--0.344) \\
\bottomrule
\end{tabular}
}
\end{table*}

\subsection{Models Trained Under Selective Labels Fail to Transport}

In the empirical analyses, we distinguish between encounters that underwent cardiac catheterization within 10 days, and untested encounters, with neither catheterization nor stress testing within 10 days (Section~\ref{sec:nightingale_data}).
Table~\ref{tab:transport} reports discrimination for
10-day revascularization among catheterized patients
and 30-day clinical outcomes among untested patients.

Models trained directly within the selectively tested population showed
substantial discrimination for revascularization among catheterized patients
but failed to preserve this ordering outside the testing pathway
(Table~\ref{tab:transport}). Across ST-MEM, ECGFounder, and ECGFlow,
catheterization-cohort AUC ranged from 0.740 to 0.770, whereas all three models
had below-chance discrimination for 30-day MACE among untested patients
(AUC 0.381--0.478). Hence, the reversal was not specific to a single ECG
architecture.

This pattern was consistent across alternative downstream
outcomes: all three models showed below-chance
discrimination for 30-day mortality, while discrimination
for troponin-positive MACE was below chance for
ECGFounder and ECGFlow and near chance for ST-MEM
(Table~\ref{tab:transport}). These endpoints capture
complementary aspects of clinical risk, including
myocardial injury and mortality, indicating that the
findings are not specific to the definition of MACE.

\subsection{Transport Depends on Target Recoverability}

\begin{figure}[!htbp]
\floatconts
{fig:target_deciles}
{\caption{
\textbf{Observed 30-day MACE risk across ECG model score deciles.}
Models trained to predict different ACS subtypes in Chongqing were applied
to untested Nightingale patients. For each model, patients were divided
into ten equally sized groups by predicted score, and observed 30-day
MACE incidence was calculated within each group. OMI shows increasing
risk, UA decreasing risk, and composite ACS a U-shaped relationship,
resulting in near-chance overall discrimination.
}}

{
\includegraphics[width=\linewidth]{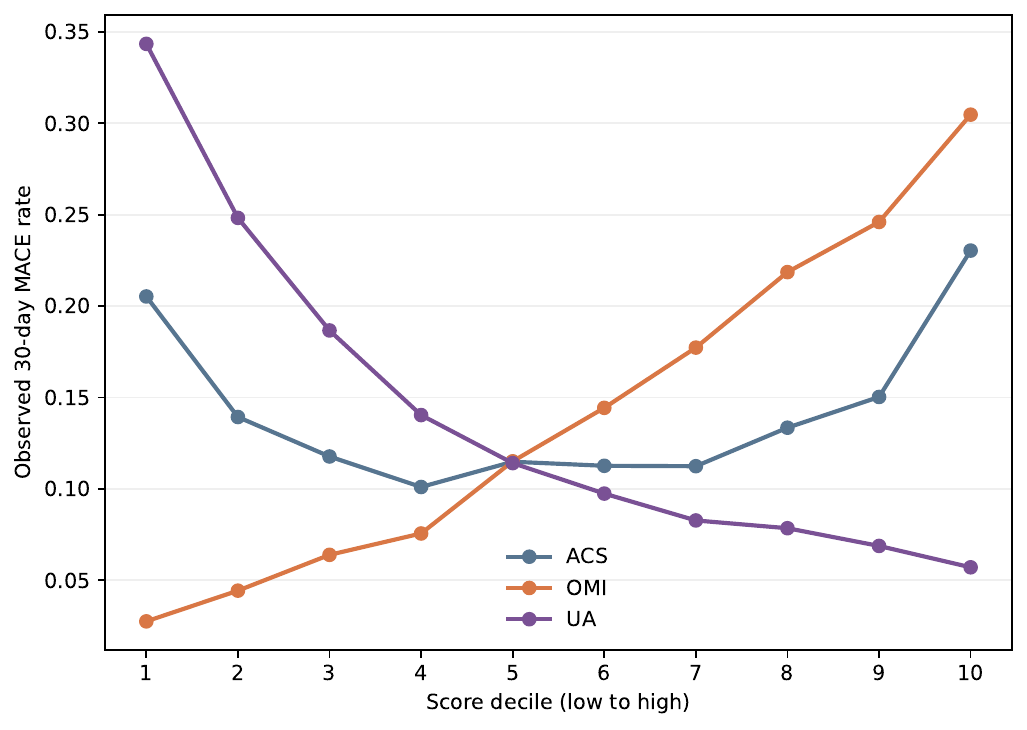}
}
\end{figure}

Target-specific models showed a qualitatively different pattern
(Table~\ref{tab:transport}). OMI, STEMI, and NSTEMI predictors retained
substantial positive discrimination for 30-day MACE in the untested
Nightingale population (AUC 0.680--0.709). In contrast, higher UA-model scores were associated with
lower observed 30-day MACE risk among untested patients
(AUC 0.310), while the broad ACS predictor was nearly
uninformative (AUC 0.517). The same pattern held for troponin-positive MACE and
30-day mortality: OMI, STEMI, and NSTEMI retained
positive discrimination, UA showed an inverse
association, and broad ACS remained near chance
(Table~\ref{tab:transport}).  These differences are consistent with internal
validation in Chongqing, where discrimination varied
substantially across clinical targets
(Appendix~\ref{app:chongqing_performance}).

The risk gradients in Figure~\ref{fig:target_deciles} illustrate this divergence. Among untested patients, 30-day MACE increased from 2.7\% to 30.5\% across OMI-score deciles, but decreased from 34.3\% to 5.7\% across UA-score deciles. The broad ACS model appears to capture both signals: ECG patterns associated with myocardial infarction and misleading associations learned from patients selected for testing despite little ECG evidence of disease. This mixture produces a U-shaped risk curve, with elevated risk at both ends of the score distribution and near-chance overall discrimination.

\section{Discussion}
\label{sec:discussion}

We identify target recoverability as an important dimension of learning under selective labels. Existing work has shown that selective testing can bias which outcomes are observed and, consequently, both model development and evaluation \citep{lakkaraju2017selective,chang2022disparate}. Our results highlight a distinct interaction: when a clinically meaningful target contains components that are poorly recoverable from the model input, conditioning on testing can make those components appear predictive within the labeled population. In a non-selective setting, an unrecoverable component should primarily contribute irreducible uncertainty. Under selective observation, however, it can instead generate a systematic shortcut and even reverse risk ranking outside the selected population. In our simulations, neither selective testing nor limited target recoverability caused the reversal on its own. It emerged only when both were present.

Prediction of Acute Coronary Syndrome provides a clinical example of this failure mode. ECG models trained directly to predict revascularization among catheterized patients discriminated well within that population but reversed their risk ordering when applied to untested patients. More importantly, transport depended strongly on the clinical target used for learning. Predictors of OMI, STEMI, and NSTEMI retained positive discrimination for subsequent adverse events. The UA predictor showed an inverse relationship with risk, while a model trained on an ACS definition combining subtypes with clear ECG manifestations and those that may lack them was nearly uninformative. The corresponding risk gradients, monotonically increasing for OMI, decreasing for UA, and U-shaped for ACS, closely mirror the behavior predicted by the theoretical framework. This interpretation is clinically plausible. Current guidelines emphasize that non-ST-elevation ACS, including UA, may occur with a normal or nonspecific ECG, and that absence of ECG evidence of ischemia does not exclude ACS \citep{rao2025acs,gulati2021chestpain}. 

The practical implication is that a clinically valid outcome is not necessarily an effective learning target under selective labels. Choosing an appropriate target requires understanding which aspects of the disease are detectable from the model's inputs and how clinical testing decisions affect the observed labels. In ACS prediction, this favors targets tied more directly to ECG-manifested pathology, such as acute occlusion, while phenotypes dependent on information outside the ECG may require multimodal inputs incorporating symptoms, biomarkers, or other clinical data. More generally, models developed from selectively observed outcomes should be evaluated whenever possible in populations outside the testing pathway, using downstream outcomes or other independent sources of information.

Researchers developing clinical machine learning models under selective labels should carefully assess whether the intended prediction target is recoverable from the available inputs. This requires understanding both the biological processes underlying the disease and the clinical information that determines which patients undergo testing. This challenge extends beyond cardiovascular disease. For example, ductal carcinoma in situ often manifests as microcalcifications on mammography, whereas invasive lobular carcinoma can be mammographically occult and detected through additional imaging or clinical findings \citep{grimm2022dcis,reig2025lobular}. A model trained on biopsy-confirmed breast cancers may therefore learn from a mixture of cases with clear mammographic manifestations and cases selected for reasons not captured in the images. Recognizing this distinction could inform the choice of prediction targets, favoring targets with clear mammographic manifestations for image-only models or incorporating additional clinical information when predicting broader cancer outcomes.

\section{Conclusion}
\label{sec:conclusion}

Selective labels can interact with target recoverability to create predictive relationships that do not transport beyond the labeled population. When clinical labels include cases that cannot be reliably predicted from the model's inputs, selective testing can create misleading associations that fail outside the tested population. Developing models that generalize requires choosing targets that reflect information available in the inputs and evaluating their performance in the populations where they will be used.




\newpage

\bibliography{ref}

\clearpage
\appendix

\section{Dataset Construction and Summary Statistics}
\label{app:data}

\subsection{Chongqing Angiography Cohort}
\label{app:chongqing_data}

We constructed the Chongqing cohort from the public dataset released by \citet{du2026acs}. We matched each metadata record to its corresponding ECG waveform and excluded recordings that did not contain the expected 5,000 samples. We defined ACS as the presence of STEMI, NSTEMI, or UA. We then divided the cohort into training, validation, and internal test sets, stratifying by ACS and OMI status and using a fixed random seed. Splits were made at the patient level to ensure that no patient appeared in more than one set. Table~\ref{tab:chongqing_summary} reports the resulting cohort sizes and label prevalences.

\begin{table}[H]
\floatconts
{tab:chongqing_summary}
{\caption{
\textbf{Overview of the Chongqing cohort by data split.}
Counts (\%) are reported per ECG; diagnostic categories may overlap.
}}
{
\centering
\scriptsize
\setlength{\tabcolsep}{3pt}
\renewcommand{\arraystretch}{1.15}
\begin{tabular}{lrrr}
\toprule
\textbf{Characteristic} &
\textbf{Train} &
\textbf{Validation} &
\textbf{Internal test} \\
\midrule
ECGs &
14,367 &
1,796 &
1,795 \\

Patients &
13,589 &
1,717 &
1,710 \\

ACS &
7,112 (49.5\%) &
889 (49.5\%) &
889 (49.5\%) \\

OMI &
920 (6.4\%) &
116 (6.5\%) &
115 (6.4\%) \\

AMI &
2,136 (14.9\%) &
279 (15.5\%) &
264 (14.7\%) \\

STEMI &
1,151 (8.0\%) &
148 (8.2\%) &
143 (8.0\%) \\

NSTEMI &
984 (6.9\%) &
130 (7.2\%) &
121 (6.7\%) \\

UA &
4,977 (34.6\%) &
611 (34.0\%) &
625 (34.8\%) \\
\bottomrule
\end{tabular}
}
\end{table}

\subsection{Nightingale BWH Emergency Department Cohort}
\label{app:nightingale_data}

We used the Nightingale Open Science Assessing Heart Attack Risk dataset \citep{anderson2023bwh}, linking ECG recordings to emergency department encounters and patient information. We divided encounters into two groups: catheterized, if patients underwent cardiac catheterization within 10 days, and untested, if they underwent neither catheterization nor stress testing during that period. 

For catheterized patients, we defined revascularization as stent placement or coronary artery bypass grafting (CABG) within 10 days. We measured 30-day major adverse cardiovascular events (MACE), troponin-positive MACE, and death in both groups. We applied the dataset's \textit{exclude\_modeling} flag, which excludes patients with certain pre-existing conditions, recent cardiac interventions, or other circumstances affecting their eligibility for testing.

\begin{table}[H]
\floatconts
{tab:nightingale_summary}
{\caption{
\textbf{Overview of the Nightingale cohort by testing group.}
Outcome counts (\%) are reported per encounter.
}}
{
\centering
\scriptsize
\setlength{\tabcolsep}{3pt}
\renewcommand{\arraystretch}{1.15}
\begin{tabular}{lrr}
\toprule
\textbf{Characteristic} &
\textbf{Catheterized} &
\textbf{Untested} \\
\midrule
ECGs &
4,818 &
102,925 \\

Encounters &
2,437 &
65,854 \\

Patients &
2,157 &
42,671 \\

Revascularization &
796 (32.7\%) &
-- \\

30-d MACE &
1,210 (49.7\%) &
8,362 (12.7\%) \\

Troponin-MACE &
1,154 (47.4\%) &
6,279 (9.5\%) \\

30-d death &
131 (5.4\%) &
3,008 (4.6\%) \\
\bottomrule
\end{tabular}
}
\end{table}

\subsection{Internal Performance of Target-Specific Models in Chongqing}
\label{app:chongqing_performance}

To characterize how well each ACS-related target can be recovered from the ECG before evaluating cross-cohort transport, we assess each target-specific model on the patient-disjoint Chongqing internal test set. The cohort contains pre-angiography 12-lead ECGs with labels for ACS, OMI, AMI, STEMI, NSTEMI, and UA \citep{du2026acs}. Table~\ref{tab:chongqing_internal} reports discrimination for each model's own training target and, for comparison, its discrimination for the composite of the different ACS subtypes. 

Discrimination for the models' own targets varied
substantially, from 0.908 for STEMI to 0.618 for UA.
When evaluated against the composite ACS outcome,
the target-specific models showed broadly similar
discrimination to their performance for revascularization
among catheterized  patients
(Table~\ref{tab:transport}) in the Nightingale Open Science Assessing Heart Attack Risk dataset, despite the difference in population. This comparison does not suggest a uniform deterioration in discrimination after cross-cohort transfer.

\begin{table}[!htbp]
\floatconts
{tab:chongqing_internal}
{\caption{
\textbf{Internal test performance of ECG models in the Chongqing cohort.}
AUCs (95\% CI) for models predicting their respective targets (own-label)
and the composite ACS label.
}}
{
\centering
\small
\setlength{\tabcolsep}{4pt}

\begin{tabular}{lcc}
\toprule

\textbf{Target} &
\shortstack{\textbf{Own-label}\\\textbf{AUC (95\% CI)}} &
\shortstack{\textbf{Composite ACS}\\\textbf{AUC (95\% CI)}} \\
\midrule
ACS &
\shortstack{0.648\\(0.623--0.676)} &
\shortstack{0.648\\(0.623--0.676)} \\
OMI &
\shortstack{0.866\\(0.835--0.896)} &
\shortstack{0.603\\(0.576--0.631)} \\
AMI &
\shortstack{0.842\\(0.815--0.868)} &
\shortstack{0.614\\(0.588--0.642)} \\
STEMI &
\shortstack{0.908\\(0.882--0.933)} &
\shortstack{0.614\\(0.588--0.641)} \\
NSTEMI &
\shortstack{0.737\\(0.692--0.779)} &
\shortstack{0.606\\(0.578--0.633)} \\
UA &
\shortstack{0.618\\(0.590--0.643)} &
\shortstack{0.462\\(0.434--0.489)} \\

\bottomrule
\end{tabular}

}
\end{table}


\subsection{Comparison with Selection-Bias Correction}
\label{app:ipw}

We also examined whether inverse-odds weighting could
prevent the ranking reversal observed in our simulations.
We used a finite-sample version of the experiment in
Section~\ref{sec:simulation}, drawing
$X\sim\mathrm{Uniform}(0,1)$ and generating mutually
exclusive outcomes $Y_R$ and $Y_U$. The unrecoverable
outcome was sampled independently of $X$ as
$Y_U\sim\mathrm{Bernoulli}(u)$, with
\[
u = \frac{0.20q}{1-0.80q}.
\]
Among observations without $Y_U$, the probability of
$Y_R$ was $0.05+0.30X$. We set $Y_R=0$ whenever
$Y_U=1$. Testing followed the same policy as in the
main simulation, depending on both $X$ and the
additional clinical signal $C$.

We compared four approaches. The first learned the
combined outcome $Y$ without weighting. The second
also predicted $Y$, but weighted observations by their
probability of being tested, estimated using $X$
alone. The third used the true testing probabilities
conditional on both $X$ and $C$, giving us an oracle
correction for selection. The fourth learned only
the recoverable outcome $Y_R$, without weighting.

For the two weighted approaches, each tested
observation received an inverse-odds weight
\[
w_i = \frac{1-e_i}{e_i},
\]
where $e_i$ is its estimated or true probability of
testing. These weights adjust the training sample
toward the untested population. For the second
approach, we estimated testing probabilities using
a model fitted to testing status and spline features
of $X$ in the full source training sample. Outcome
labels from untested observations were not used.
For the oracle approach, we used the known testing
probabilities from the simulation. 

All four outcome models used logistic regression
with the same cubic spline basis of $X$, fitted
only among tested observations. The basis had six
fixed knots at $0,0.2,0.4,0.6,0.8,1$ and seven
retained features. Every model used only $X$ at prediction
time, including the oracle-weighted model, which
used $C$ solely to calculate training weights.

We repeated the experiment ten times, independently
generating 30,000 training and 30,000 evaluation
observations each time. Models were trained on the
tested observations and evaluated against the same
combined outcome $Y$ among untested observations.
At $\lambda=1$ and $q=0.70$, each repetition
contained an average of 18,784 tested training
observations and 11,249 untested evaluation
observations. Table~\ref{tab:ipw} reports the mean AUROC, Brier score, and log loss across ten independent repetitions, with 95\% confidence intervals.

\begin{table*}[!tbp]
\floatconts
{tab:ipw}
{\caption{
\textbf{Simulation of inverse probability weighting under selective testing.}
Performance of four training approaches ($\lambda=1$, $q=0.70$),
evaluated against the composite outcome $Y$ among untested observations.
Values are means and 95\% confidence intervals across ten repetitions.
Lower Brier scores and log loss indicate better calibration and
probability predictions.
}}
{
\centering
\small
\setlength{\tabcolsep}{5pt}
\begin{tabular}{lccc}
\toprule
\textbf{Training approach} &
\textbf{AUROC (95\% CI)} &
\textbf{Brier score (95\% CI)} &
\textbf{Log loss (95\% CI)} \\
\midrule
Broad-label learning &
0.378 (0.373--0.383) &
0.325 (0.322--0.329) &
0.851 (0.843--0.859) \\
Observed-input weighting &
0.378 (0.373--0.383) &
0.325 (0.322--0.329) &
0.851 (0.843--0.859) \\
Oracle weighting &
0.638 (0.632--0.644) &
0.183 (0.182--0.185) &
0.551 (0.547--0.554) \\
Recoverable-label learning &
0.638 (0.632--0.644) &
0.214 (0.212--0.215) &
0.692 (0.683--0.702) \\
\bottomrule
\end{tabular}
}
\end{table*}

Weighting based on $X$ alone made no difference.
Discrimination and probability accuracy were
unchanged from unweighted learning. The missing
information mattered. Because testing also depended
on $C$, the observed-input weights could not correct
the associations created by selection. Oracle
weighting, which accounted for $C$, restored
positive discrimination (AUROC 0.638) and
substantially improved performance.

Training on $Y_R$ achieved the same AUROC of 0.638,
but its probability predictions were less accurate
for the combined outcome $Y$. This is expected:
the model estimated the probability of $Y_R$,
which excludes the unrecoverable cases contributing
to $Y$.

Both approaches can therefore recover useful risk
rankings, but they require different information.
Correcting selection requires knowledge of the
factors driving testing. Learning a recoverable
target requires labels that distinguish the
underlying outcome types. Our simulation shows
how these approaches behave under a known testing
policy.

\subsection{MACE Performance of Chongqing Models}
\label{app:tested_mace}

We also report discrimination of the Chongqing-trained
models for 30-day MACE in catheterized and untested
Nightingale patients (Table~\ref{tab:tested_mace}).
MACE may reflect different clinical processes in these
two groups. Catheterized patients undergo further diagnostic
evaluation and may receive treatment, so subsequent events
may reflect disease severity or occur despite treatment.
Among untested patients, MACE may also capture consequences
of missed ACS. The results should therefore be interpreted
in the context of these differences in clinical care.

\begin{table}[H]
\floatconts
{tab:tested_mace}
{\caption{
\textbf{ECG model performance for 30-day MACE in catheterized and untested patients.}
AUCs (95\% CI) for models trained on different ACS targets in the Chongqing cohort and evaluated in the Nightingale cohort.
}}

{
\centering
\small
\setlength{\tabcolsep}{4pt}
\begin{tabular}{lcc}
\toprule
\textbf{Target} &
\shortstack{\textbf{Catheterized}\\\textbf{AUC (95\% CI)}} &
\shortstack{\textbf{Untested}\\\textbf{AUC (95\% CI)}} \\
\midrule
ACS &
\shortstack{0.573\\(0.548--0.598)} &
\shortstack{0.517\\(0.492--0.535)} \\
OMI &
\shortstack{0.633\\(0.608--0.658)} &
\shortstack{0.709\\(0.702--0.716)} \\
AMI &
\shortstack{0.629\\(0.605--0.653)} &
\shortstack{0.687\\(0.676--0.696)} \\
STEMI &
\shortstack{0.633\\(0.608--0.657)} &
\shortstack{0.693\\(0.683--0.703)} \\
NSTEMI &
\shortstack{0.613\\(0.589--0.637)} &
\shortstack{0.680\\(0.672--0.688)} \\
UA &
\shortstack{0.402\\(0.376--0.425)} &
\shortstack{0.310\\(0.296--0.321)} \\
\bottomrule
\end{tabular}
}
\end{table}

\end{document}